\pdfoutput=1
\documentclass{llncs}
\usepackage{amssymb}
\usepackage{graphicx}
\usepackage[misc]{ifsym}
\usepackage{amsmath}
\usepackage{lineno,hyperref}
\usepackage{subfig}
\usepackage{stmaryrd}
\usepackage{url}
\usepackage{color}
\newcommand{\norm}[1]{\left\lVert#1\right\rVert}
\urldef{\mailsa}\path|{yifan_liu|
\urldef{\mailsb}\path|zcqin}@buaa.edu.cn|
\urldef{\mailsc}\path|{zb.luo|
\urldef{\mailsd}\path|hua00.wang}@samsung.com|

\begin{document}
\title{Auto-painter: Cartoon Image Generation from Sketch by Using Conditional Generative Adversarial Networks}
\titlerunning{Auto-painter}

\author{Yifan Liu$^{1,2}$ \and Zengchang Qin$^{1}$ \and Zhenbo Luo$^{2}$ \and Hua Wang $^{2}$  }
\authorrunning{Liu, Qin, Luo and Wang}

\institute{$^1$Intelligent Computing and Machine Learning Lab, School of ASEE\\
Beihang University, Beijing 100191, China\\
$^2$ Samsung R{\&}D Institute China Beijing\\
 18F TaiTangGong Plaza, Beijing, 100028, China\\
\mailsa,
\mailsb,
\mailsc,
\mailsd\\
}

\maketitle

\begin{abstract}

Recently, realistic image generation using deep neural networks has become a hot topic in machine learning and
computer vision. Images can be generated at the pixel level by learning from a large collection of images.
Learning to generate colorful cartoon images from black-and-white sketches is not only an interesting
research problem, but also a potential application in digital entertainment.
In this paper, we investigate the sketch-to-image synthesis problem by using conditional generative adversarial networks
(cGAN).
We propose the auto-painter model which can automatically generate compatible colors for a sketch.
The new model is not only capable of painting hand-draw sketch with proper colors, but also allowing
users to indicate preferred colors.
Experimental results on two sketch datasets show that the auto-painter performs better that existing image-to-image methods. 

\end{abstract}

\section{Introduction}
\label{sec:intro}

Human beings possess a great cognitive capability of comprehend black-and-white cartoon sketches.
Our mind can create realistic colorful images when see black-and-white cartoons.
It needs a great artistic talent to choose appropriate colors, also with proper changes in light and shade to create vivid cartoon images.
It is not easy for ordinary people to do so. 
How to automatically paint the sketch into colorful cartoon images is a useful application for digital entertainment.
In this work, we are interested in solving this problem by employing deep neural networks with designed constraints
to transfer the line draft into specific cartoon style.
Practically, the new model can make up for ordinary people's artistic talent and even inspire the artists to create new images.
Ideally, readers may have the freedom to choose to generate different style of cartoons
based on their own tastes of colors.

Cartoon image generation from sketches can be regarded as an image synthesis problem. Previously, many non-parametric models were proposed \cite{Hays2007Scene,Barnes2009PatchMatch,Freeman2002Example} by matching the sketch to a database of existing image fragments. Recently, numerous image synthesis methods based on deep neural networks have emerged \cite{Salakhutdinov2009Deep,Lee2009Convolutional,Goodfellow2014Generative,Radford2015Unsupervised}.
These methods can generate detailed images of the real-world, such as faces, bedrooms, chairs and handwritten numbers. As the photo-realistic images are full of sharp details, the results may suffer from being blurry \cite{Kingma2014Auto}, noisy \cite{Goodfellow2014Generative} and objects being wobbly \cite{Denton2015Deep}.
What's more, the outputs of the network are hard to be controlled because the generator samples from a random vector in low dimension and the model has too much freedom. Several recent approaches have explored the applicability of controllable deep synthesis in different applications, for instance, the super resolution problem \cite{Johnson2016Perceptual}, semantic labels tagging for objects \cite{Dong2017Unsupervised}, grayscale image colorization \cite{Zhang2016Colorful}. 
Especially the sketch-to-image problem \cite{Sangkloy2016Scribbler}, the control signals are relatively sparse, more ill-posed than colorization problem based on grayscale, it requires a model to synthesize image details beyond what is contained in the input. The network should learn the high frequency textures for the details of the scene elements as well as high-level image styles.

\begin{figure}[!h]
\centering
\includegraphics[width=\textwidth]{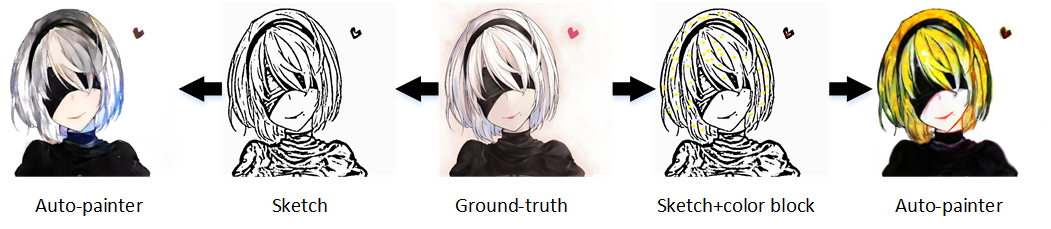}
\caption{Examples of painted cartoon images generated from a sketch by auto-painter. }
\label{fig:firstFig}
\end{figure}

In this paper, we propose a learning model called \emph{auto-painter}  that can automatically generate painted cartoon images from a sketch based on conditional Generative Adversarial Networks (cGANs).
The cartoon images have more artistic color collocations (e.g. green hair, purple eyes) that may require more constraints in modeling.
Constraints including \emph{total variance loss}, \emph{pixel loss} and \emph{feature loss} are used in training the generator
in order to generate more artistic color collocations.
We also introduce the color control for the auto-painter based on \cite{Sangkloy2016Scribbler}
to allow users paint their favorite colors. Figure \ref{fig:firstFig} shows an example of generated cartoon images from a
sketch, the results of auto-painter (with and without color control) are compared to the ground truth in the middle.


\section{Related Work}
\label{sec:rela}
{\bf Generative Adversarial Networks}
Generative adversarial networks (GANs) were recently regarded as a breakthrough in machine learning \cite{Goodfellow2014Generative,Radford2015Unsupervised}, it consists of two `adversarial' modules: a generative module $G$ that captures the data distribution, and a discriminative module $D$ that estimates the probability that a sample came from the training data rather than $G$. Both $G$ and $D$ could be deep neural networks.
In image synthesis with GANs, the generator attempts to produce a realistic image from an input random vector to fool the simultaneously adversarial trained discriminator, which tries to distinguish whether its input image is from the training set or the generated set. It corresponds to a minimax two-player game.
The generator has benefited from convolutional decoder networks, it can go back to the work using deep convolutional decoder networks to generate realistic images.
\\
{\bf Conditional GANs}
GANs are unconditioned generative models that learn a mapping from random noise vector $z$ to output $y$:
$G : z \rightarrow y$. In contrast, conditional GANs (cGANs) learn a mapping from observed input
$x$ and random noise vector $z$, to $y$: $G : \{x, z\} \rightarrow y$ \cite{Mirza2014Conditional}.
Several works studied different cGAN where the generator is conditioned on inputs such as text \cite{Reed2016Generative}, labels \cite{Eigen2015Predicting} and other forms of the images \cite{Isola2016Image} to generate `fake' images.
The discriminator needs to distinguish the authentic and fake pairs of images \cite{Reed2016Generative}.
In this paper, both of generator and discriminator are conditioned on the input sketch in order to get a better
supervised performance. \\
{\bf Image-to-image} The `pix2pix' method proposed in \cite{Isola2016Image} is a `U-net' architecture \cite{Ronneberger2015U} to deal with general image-to-image transfer which allows the decoder to be conditioned on encoder layers to get more information. They investigated different kinds of image-to-image transfer tasks that include
transforming the image of daylight to night, produce city images from map, and even synthesize shoes and handbags from designer's sketches.
Sangkloy \emph{et al.}\cite{Sangkloy2016Scribbler}
proposed the `scribbler'  by allowing users to scribble over the sketch to indicate preferred color for objects.
In our model, we also introduce the function of interactive color control to generate different style of cartoon images.


\section{Method}
\label{sec:meth}

Auto-painter is a supervised learning model, given a black-and-white sketch, the model can generate
a painted colorful image based on given sketch-image pairs in the training data.
We employ a feed-forward deep neural network as generator to get a quick response in test.
The generator takes the sketch as input and outputs a colorful cartoon image of the same resolution at pixel-level.

\subsection{Network Structure}
\label{se:ns}
\begin{figure*}[!h]
\centering
\hbox{
\includegraphics[width=0.48\textwidth]{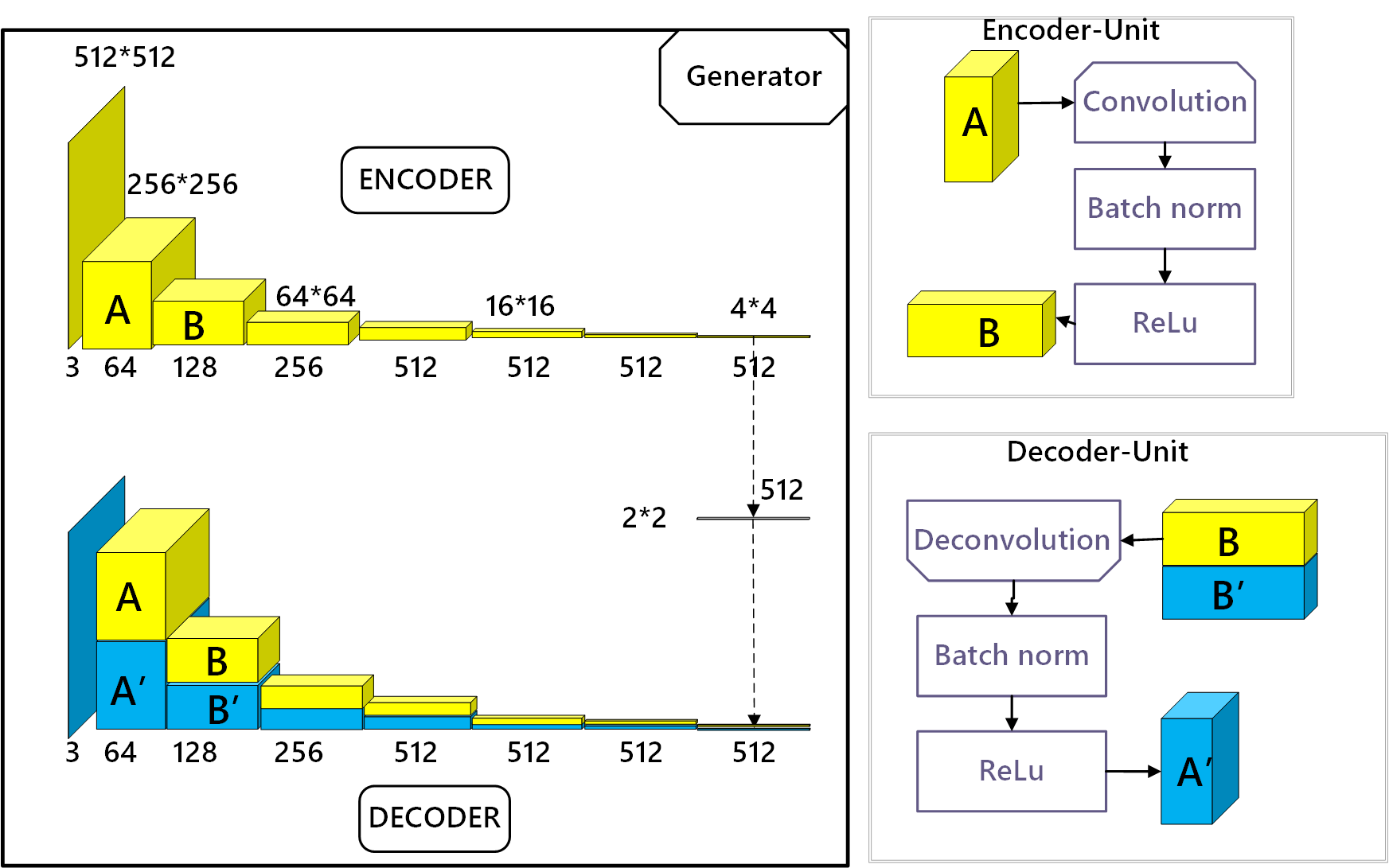}
\includegraphics[width=0.52\textwidth]{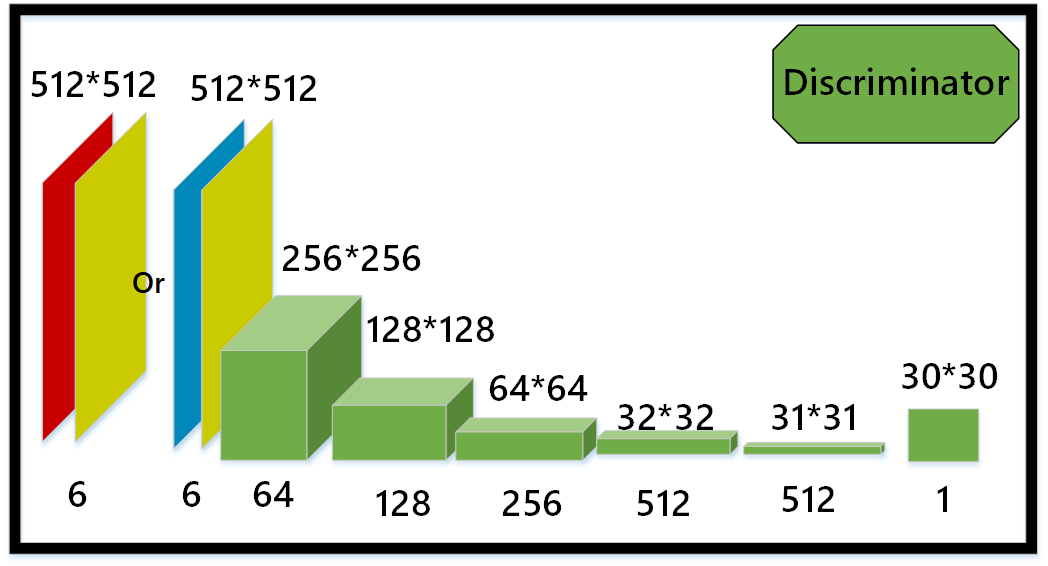}}
\caption{Left-hand side: the U-net structure of generator: the yellow blocks represent layers in encoder, and blue blocks are decoder layers.
In each layer of decoding, the corresponding layers of encoder are concatenated to current layer to decode the next layer.
E.g., $B$ and $B'$ are concatenated to obtain $A'$ through deconvolution. The right-hand side figure is the patchGAN discriminator: input of discriminator is either the pair of sketch (yellow block) and real target image (red block), or the pair of sketch and generated image (blue block).
}
\label{fig:gen}
\end{figure*}

Many previous solutions \cite{Dong2014Learning,Sangkloy2016Scribbler,Wang2016Generative} used a pure encoder-decoder network. The input goes through a series of down sampling steps to a lower dimension, then gets some non-liner transformation with a fully connected layer, and finally gets up sampling to the present output size.
Such a structure may cause information loss when passing through layers.
Especially in the sketch-to-image problem, we need to keep the edges as the most important information from the input
to ensure the quality of the output image.
Instead of the encoder-decoder structure, we employ the `U-net' \cite{Ronneberger2015U}, by concatenating
layers in encoder to the corresponding layers of decoder.
As we can see from Figure \ref{fig:gen}, in order to decoding out low-level information of sketch,
we concatenate encoder layer $A$ to the decoding layer $A'$ for generating the final colorful cartoon image,
where $A$ contains the sketch edge information and $A'$ mainly contains trained color
painting information.

The discriminator only has the encoder units comparing to the generator, it aims to classify whether the input sketch-image
pair is `real' or `fake'.  The network is trained to maximize the classification accuracy.
As we can see from Figure \ref{fig:gen} (right), the output of the discriminator is a matrix of probability, in which each element gives the
probability of being real for a pair of correspond patches sampled using Markov random field or PatchGAN \cite{Isola2016Image}.
In our research, we take the patch size of $70 \times 70$ and output a probability matrix of $30 \times 30$.
The convolutional layers between the input and the output extract the high-level features of the input pairs.

\subsection{Loss Function}
\label{se:LF}
The task of auto-painter is similar to colorization \cite{Iizuka2016Let} but much harder, because a gray scale image has more information than a simple line sketch. Facing this ill-posed problem, we may need more constraints for better performance.
The cGANs learn a mapping from observed $x$ and random noise vector $z$ to the output image $y$. The objective (loss) function of general cGANs can be expressed by:
\begin{equation}
\label{equ:gan}
\begin{split}
\mathop {{\rm{min}}}\limits_G \mathop {\max }\limits_D V(G,D) = &{\mathbb{E}_{x,y\thicksim{p_{data}}(x,y)}}[\log D(x,y)]+\\ &{\mathbb{E}_{x\thicksim{p_{data}}(x),z\thicksim{p_{data}}(z)}}[\log (1 - D(x,G(x,z)))]
\end{split}
\end{equation}
where the generator $G$ tries to minimize the objective function while an adversarial $D$ tries to maximize it.
As for the auto-painter, we update the weights of the encoder-unit in discriminator to maximize $V(G,D)$, while
$x$ is the input sketch, $y$ is the target (the colored cartoon image) and $G(x,z)$ is the generated image.
The discriminator $D$ outputs the probability to classify the `real' and `fake' input pair as mentioned in Section \ref{se:ns}.

The generator is simultaneous trained to minimize the generative loss:
\begin{equation}
\label{equ:G}
L_G = {\mathbb{E}_{x\thicksim{p_{data}}(x),z\thicksim{p_{data}}(z)}}[\log (1 - D(x,G(x,z)))]
\end{equation}
where we provide a Gaussian noise $z$ as an input of generator, in addition to $x$. The adversarial training encourages more variations and vividness in generated images. Previous approaches of conditional GANs \cite{Isola2016Image,Pathak2016Context}  have found it beneficial to mix the GAN objective with a more traditional loss functions. 
In this paper, we also use the $L_1$ distance to describe the pixel-level loss  ${L_p}$ in our model:
\begin{equation}
\label{equ:p}
{L_p} ={\mathbb{E}_{x,y\thicksim{p_{data(x,y)},z\thicksim{p}_{_{data}}(z)}}} \left[\norm{y - G(x,z)}_1 \right]
\end{equation}
${L_p}$ represents for the difference between the generate image and the ground truth at pixel level.
Except for the pixel loss,
we also employ a pre-trained Visual Geometry Group net (VGGnet) \cite{Simonyan2014Very} to extract high-level information of the image. The feature loss ${L_f}$ is defined as the $L_2$ distance in a feature space:
\begin{equation}
\label{equ:f}
{L_f} = {\mathbb{E}_{x,y\thicksim{p_{data(x,y)},z\thicksim{p_{data}}(z)}}} \left[\norm{{\phi _j}(y) - {\phi _j}(G(x,z))}_2\right]
\end{equation}
${\phi _j}$ is the activation function of the $j$th layer of the 16-layer VGG network (VGG16) pre-trained on the ImageNet dataset \cite{Russakovsky2015ImageNet}. In this paper, through the trail study, we choose the outputs of the fourth layer of VGG16 ($j=4$).
In order to avoid a color mutation in the output, we also add a total variation loss ${L_{tv}}$. For the ${L_{tv}}$ can constrain the pixel changes in the generated results and encourages smoothness \cite{Johnson2016Perceptual}.
\begin{equation}
{L_{tv}} = \sqrt {{{({y_{i + 1,j}} - {y_{i,j}})}^2} + {{({y_{i,j + 1}} - {y_{i,j}})}^2}}
\end{equation}
Finally, the objective function is defined as the following:
\begin{equation}
\label{equ:all}
L = {w_p}{L_P} + {w_f}{L_f} + {w_G}{L_G} + {w_{tv}}{L_{tv}}
\end{equation}
where ${w_p}$, ${w_f}$, ${w_G}$ and ${w_{tv}}$ are the weights for loss functions we discussed above.
We adjust them to control the importance of each part. This loss function ensures that the generated image
contains both pixel level details of sketch as well as high-level information of painted colors. By minimizing this
objective function, we can learn complex correlations between sketch and its associated painting style.

\section{Experimental Studies}
\subsection{Datasets}
\label{sec:dat}
\begin{figure*}
\centering
\includegraphics[width=0.9\textwidth]{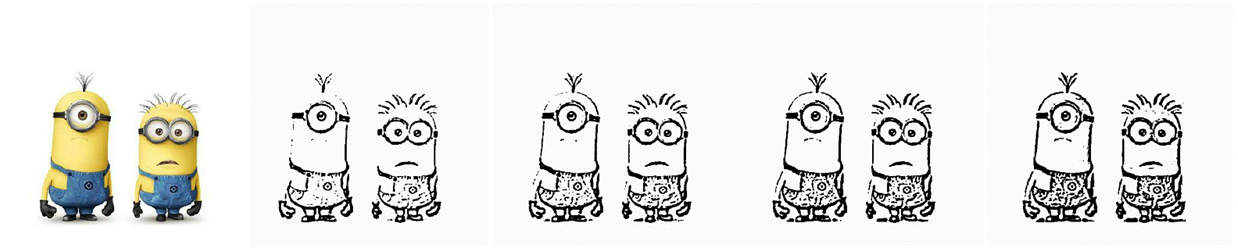}
\caption{Generated black-white sketches using XDoG filter \cite{Winnem2012XDoG}. From left to the right: the parameter $\gamma$ is set to 0.96, 0.97, 0.98, 0.99 in order to obtain sketches with different level of high-frequency information.}
\label{fig:dog}
\end{figure*}

In order to train the auto-painter model, we collect a large number of cartoon pictures from the Internet with a crawler.
Previous researches \cite{Sangkloy2016Scribbler,Zhang2016Colorful} prefer lower resolution for the synthetic images.
To be more practical as most cartoons are in higher resolution, our training images are all resized to $512\times512$.
This will increase the difficulty of training, but we deepen the neural network and use more constraints to stabilize the training.
It is not easy to find paired sketch-image cartoons for training, we employ the boundary detection filter XDoG \cite{Winnem2012XDoG} to extract sketches. By adjusting parameters $\gamma$ of the XDoG, we can obtain sketches with different levels of details (e.g, see Figure \ref{fig:dog}). We include all four types of sketches in training.
All the data will be available to public at the final version of this paper.
After preprocessing (resize, truncate to squared images and extract sketches), two cartoon datasets are obtained: \emph{Minions} and \emph{Japanimation}. \emph{Minions} contains 1100 pictures of different colored minions. \emph{Japanimation} contains 60000 pictures of Japanese anime, and most of them are characters. We take 90 percent of data for training and 10 percent for test.

\subsection{Evaluation Metrics}
It is well-known that evaluating the quality of a synthesized image is an open and difficult problem  \cite{Zhang2016Colorful}. The traditional metrics such as per-pixel mean-squared error can not assess high-level features of the output image.
In order to evaluate the visual quality of the auto-painter and compare to the pix2pix model, we design a test called `like vs dislike'
inspired by the `real vs fake' test for evaluating realistic image synthesis models in \cite{Isola2016Image,Zhang2016Colorful}.
Because the ultimate goal of the auto-painter is to produce favorite cartoon images.
Volunteers were presented with a series of painted images generated by different algorithms and asked to choose the best one and the worst one.
In this research, we design this evaluation metrics mainly to analyze the effect of the different parts of the loss function. Unlike \cite{Zhang2016Colorful}, we post four pictures simultaneously for two seconds and complete the evaluation of all four algorithms instead of evaluating them one by one.
This allows the participants to conclude after a more comprehensive comparison. No feedbacks are given during the test because
all images are generated `fake' ones, and there is no prejudice on
auto-painter' s results.  In our paper, 55 volunteers were asked to take part in the evaluation.

In order to quantify the evaluation, we define the \emph{popularity index} as the following: for the $i^{th}$ picture generated by the $j^{th}$ algorithm, the popularity $pop_{ij}$ is define by:
${pop_{ij}} = \log (\frac{{n^{like}_{ij} + c}}{{n^{dislike}_{ij} + c}})$, where $i \in [1,40]$, $j \in [1,4]$, $c=1$ in our test and ${n^{like}_{ij}}({n^{dislike}_{ij}})$ represent the number of subjects who choose `like(dislike)' on this ${picture_{ij}}$. $c$ is set for smoothness.
Therefore, the popularity of ${j^{th}}$ algorithm is:
${pop_{j}} = \log (\frac{{\sum\limits_i {n_{ij}^{like}}  + c}}{{\sum\limits_i {n_{ij}^{dislike}}  + c}})$
We consider both ${pop_{j}}$ and ${pop_{ij}}$ because for different kinds of sketches, a specific algorithm may output different quality images. The variance of ${pop_{ij}}$ represents for the stability of an algorithm $j$.

\subsection{Experimental Results}

\begin{figure}[h]
\centering
\includegraphics[width=0.9\textwidth]{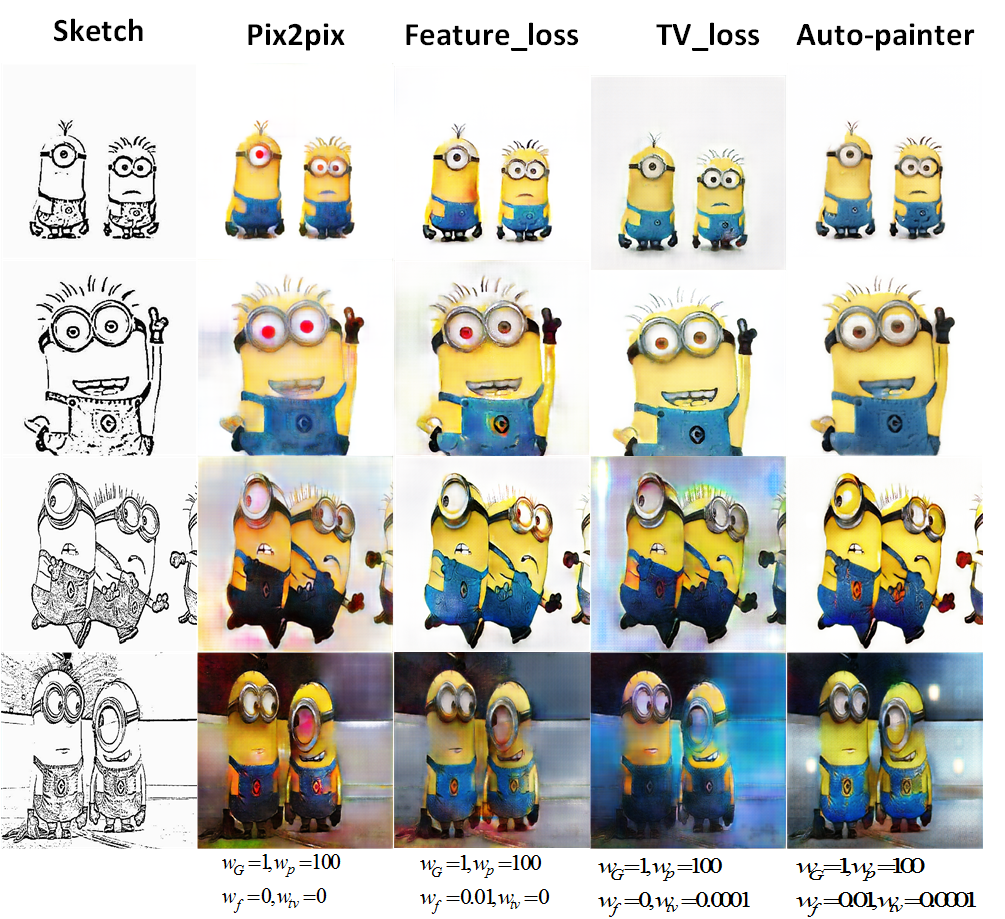}
\caption{Results of auto-painter and pix2pix with different parameter settings.}
\label{fig:com}
\end{figure}

In order to verify the importance of constraints ($L_f$ and $L_{tv}$) introduced in Section \ref{se:LF}, we make ablation studies to isolate the effect of each term.
The Pix2pix model which has no feature loss and total variation (tv) loss is used as the baseline method
to compare to the results of the auto-painter.
All the parameters of training details are the same including the random seeds, learning rate, epoches and the batch size. The only difference are the objective function and the detail weights are shown in Figure \ref{fig:com}. It also gives qualitative results of these variations on the Minions.
Without tv loss (${w_{tv}}=0$), the result images tend to dissolve to background and look messy.
If we set ${w_{f}}=0$, which means we do not consider the feature loss, the details of result image become blurry.
The relative importance of ${L_{p}}$ and ${L_{G}}$ have already been discussed in \cite{Isola2016Image}.
Considering both terms together may produce quality results.
Table \ref{tab:result} show result comparisons of the pix2pix model \cite{Isola2016Image}, pix2pix $+$tv loss,
pix2pix $+$feature loss and the auto-painter.
From the table, we can see that auto-painter is the most popular algorithm among all candidates while pix2pix is the worst.
Adding a tv loss improves the popularity but decreases the stability as the variance is larger.
Adding feature loss improves the ability of the system to adapt for diversity of sketches.
However, combine the tv loss and the feature loss together we will have the largest popularity and the smallest variance.
In summary, auto-painter considers all these constraints to fit all sorts of sketches (with or without background,
hand-draw or modified) and it can achieve
the best performance.
\begin{table}
\begin{center}
\begin{tabular}{|l|c|c|c|c|}
\hline
Method& pix2pix \cite{Isola2016Image} & tv loss & feature loss & auto-painter \\
\hline\hline
${n^{like}}$ &249&304&687&960\\
${n^{dislike}}$ &1147&698&219&136\\
${pop_{j}}$ &-1.524 &-0.829 &1.140 &1.948\\
${variance(pop_{ij})}$ &1.319 &1.519 &1.110 &0.888 \\
${mean(pop_{ij})}$&-1.549 &-0.675 &1.227 &1.873\\
\hline
\end{tabular}
\end{center}
\caption{Results of \emph{`like vs dislike'} test with 55 volunteers. }
\label{tab:result}
\end{table}

\begin{figure}
\centering
\hbox{
\includegraphics[width=0.5\textwidth]{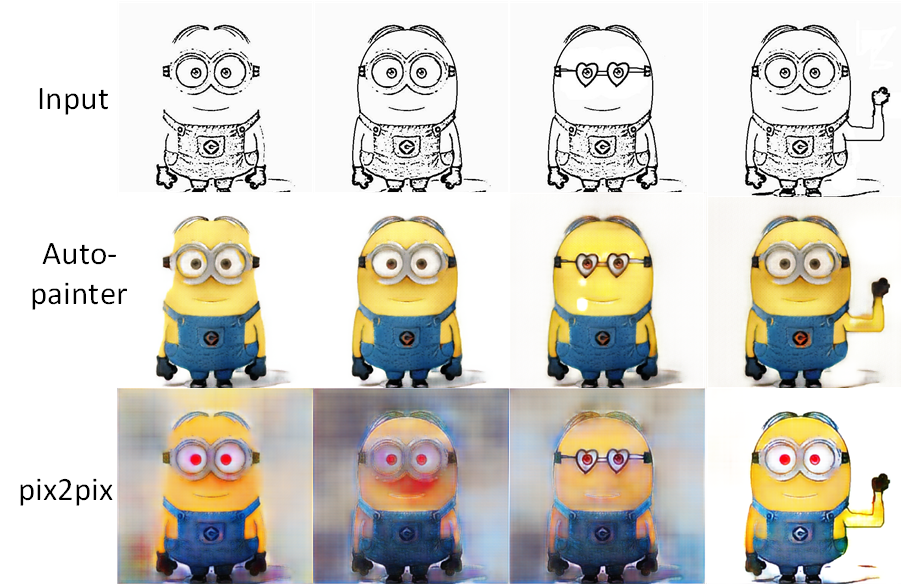}
\includegraphics[width=0.5\textwidth]{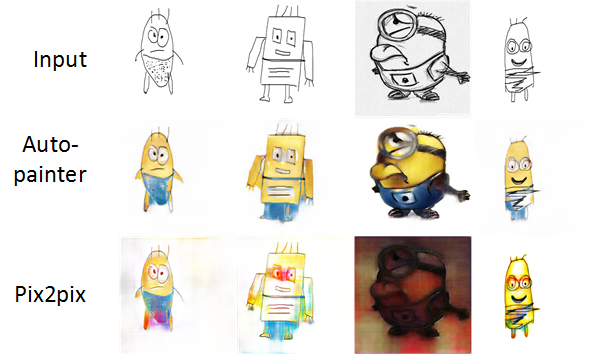}}
\caption{Cartoon images generated from modified and amateur hand-draw sketches. }
\label{fig:cha}
\end{figure}

Given an input image of resolution $512 \times 512$, the auto-painter can generate  a painted colorful output image within 1 second, which enables instant feedback to design an interactive image editing tools.
The auto-painter trained on the Minions provides a tool for users to design virtual images of `minions style'. As it shown in
the left-hand side of Figure \ref{fig:cha}, based on the given initial sketch, one can modify the original sketch, the glass or the gesture, etc. It is interesting that even on the simple hand draw sketches, the auto-painter can result in `minions style' images (right-hand side of Figure \ref{fig:cha}).

\subsection{Color Controlled Sketch Synthesis}

When a gray-scale sketch is transferred to a real-colored photo, the color scheme is relatively fixed because it shows the
color of the real-world, such as grass is green, sky is blue and sand is golden. There may be a mass of alternative color scheme in a complex training set.
Especially in the cartoon image dataset, the color of a girl's eyes may be green or purple, which is uncommon in the real world.
If we take the example of the minions, the model can learn the correlation between the minion shapes and the yellow color.
Such correlations are too hard to learn on the dataset like \emph{Japanimation}. When observing a complex cartoon sketch, different users may prefer different color schemes. But the generator trained with black-and-white sketch can only choose a specific color schemes.  So we train a color control model to satisfy different users aesthetic needs.
In order to train the auto-painter to recognize color control signals, we can add color block to sketches. We blur the ground-truth image with a Gaussian filter and sample a random number of points at random start locations and grow them to become a color block along the diagonal. It is obvious that each particular color block should not across different color regions , so we put a constraint when growing the block that if the difference between the mean color of the current block and the next sample block exceeds a threshold, we stop the growing.

\begin{figure}
\centering
\includegraphics[width=1\textwidth]{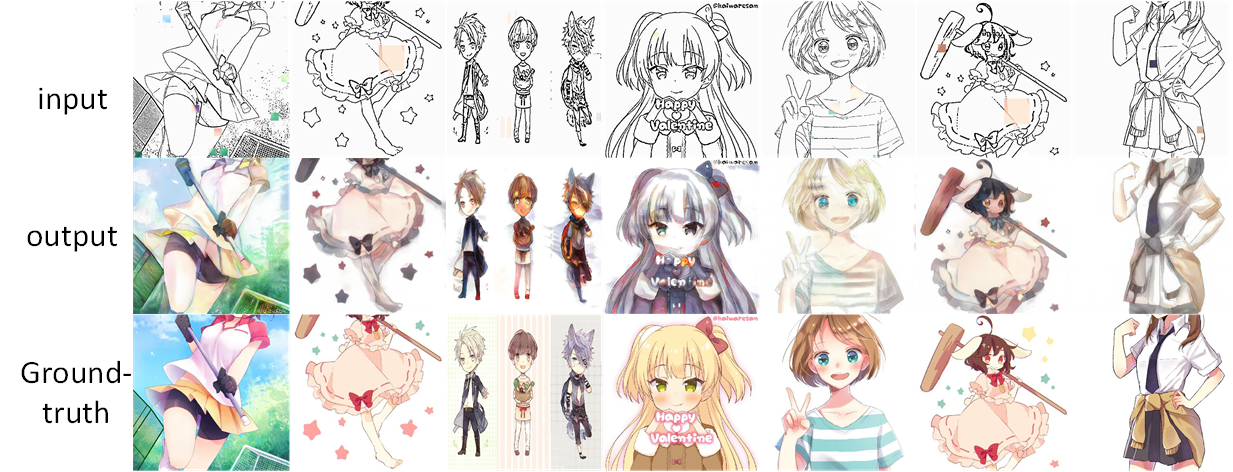}
\caption{Results on the \emph{Japanimation}. The color blocks are generated by random sampling. }
\label{fig:ct}
\end{figure}

\begin{figure}[!h]
\centering
\includegraphics[width=0.95\textwidth]{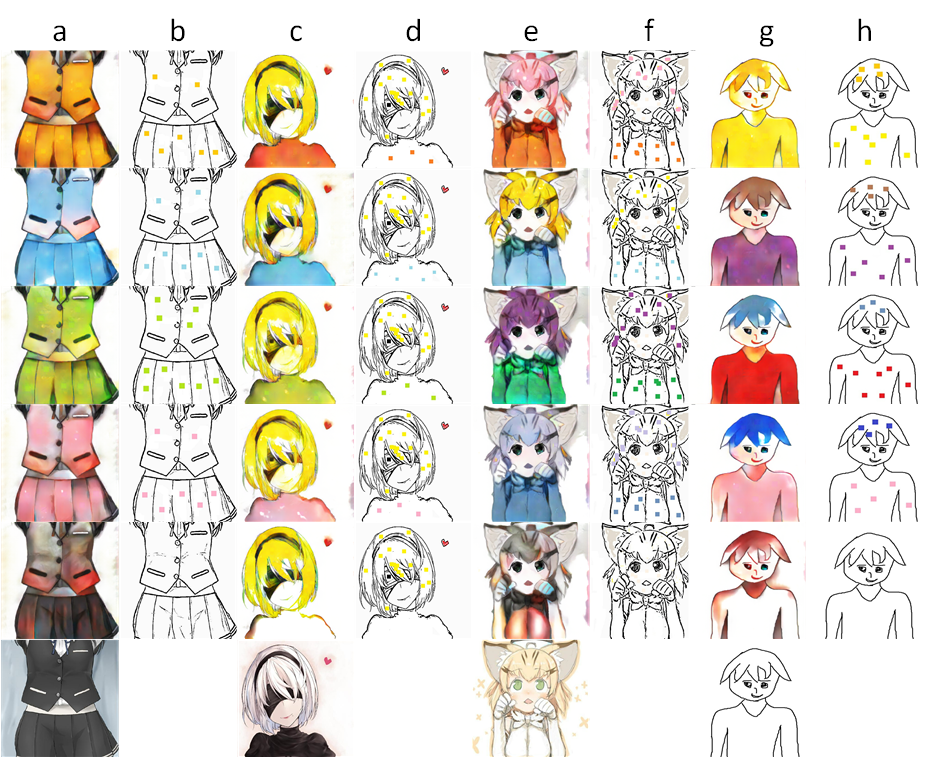}
\caption{Examples of painted sketch by auto-painter with color control: (a), (c), (e), (g) are generated cartoon
 images with corresponding inputs with indicated color (b), (d), (f), (h).
The last row is the ground truth from the training data.}
\label{fig:c}
\end{figure}

In the experiment on the \emph{Japanimation} data set, we need the model to generate more colorful images comparing to
the Minions data.
Follow the description above, we generate sketches with color control blocks of \emph{Japanimation}.
Figure \ref{fig:ct} shows the results of reconstructing \emph{Japanimation} based on synthesis color control blocks. We can see that for the complicated task, the auto-painter identify the edges of the images successfully and make feasible color schemes based on the color block.
The color of the control block is rendered around in a particular region naturally. We also test the model with user input and the colors of the blocks deviate a lot from the colors in the ground-truth image (see Figure \ref{fig:c}). Nevertheless, the auto-painter is able to paint the color in object boundaries and distinguish different parts of the sketch.

\section{Conclusion}
In this paper, we proposed the auto-painter model to solve the sketch-to-image problem.
Our approach was based on conditional GAN with `U-net' structure that allows the output image has
both low level information of sketch as well as learned high-level color information.
We introduce more constraints based on the pix2pix model to obtain better painting performance.
We also trained the auto-painter to adapt to color control, so that our network can adapt the synthesis results to satisfy users with different color taste.
Empirical results show that the auto-painter can generate high-quality cartoon images on two given datasets, and it has the best popularity measure based on a subjective test with volunteers.
Despite the promising results, our current system suffers from the difficulties of adjusting parameters just like other deep learning models. Complex network structure may result in low training speed. For example, on our \emph{Japanimation} dataset, it takes 2 to 3 days to get fairly good results with a single Tesla K80 GPU. In the future work, we will focus on improving the
system performance as well the speed in order to build an
interactive toolkit for users.
\bibliographystyle{splncs03}
\bibliography{egbib}

\begin{thebibliography}{10}
\providecommand{\url}[1]{\texttt{#1}}
\providecommand{\urlprefix}{URL }

\bibitem{Barnes2009PatchMatch}
Barnes, C., Shechtman, E., Finkelstein, A., Dan, B.G.: Patchmatch:a randomized
  correspondence algorithm for structural image editing. ACM Trans. Graph.
  28(3),  1--11 (2009)

\bibitem{Denton2015Deep}
Denton, E.L., Chintala, S., Szlam, A., Fergus, R.: Deep generative image models
  using a laplacian pyramid of adversarial networks. In: NIPS (2015)

\bibitem{Dong2014Learning}
Dong, C., Chen, C.L., He, K., Tang, X.: Learning a deep convolutional network
  for image super-resolution. ECCV  8692,  184--199 (2014)

\bibitem{Dong2017Unsupervised}
Dong, H., Neekhara, P., Wu, C., Guo, Y.: Unsupervised image-to-image
  translation with generative adversarial networks. CoRR  abs/1701.02676 (2017)

\bibitem{Eigen2015Predicting}
Eigen, D., Fergus, R.: Predicting depth, surface normals and semantic labels
  with a common multi-scale convolutional architecture. In: IEEE International
  Conference on Computer Vision. pp. 2650--2658 (2015)

\bibitem{Freeman2002Example}
Freeman, W.T., Jones, T.R., Pasztor, E.C.: Example-based super-resolution. IEEE
  Computer Graphics {\&} Applications  22(2),  56--65 (2002)

\bibitem{Goodfellow2014Generative}
Goodfellow, I.J., Pougetabadie, J., Mirza, M., Xu, B., Wardefarley, D., Ozair,
  S., Courville, A., Bengio, Y., Ghahramani, Z., Welling, M.: Generative
  adversarial nets. Advances in Neural Information Processing Systems  3,
  2672--2680 (2014)

\bibitem{Hays2007Scene}
Hays, J., Efros, A.A.: Scene completion using millions of photographs. Commun.
  ACM  51,  87--94 (2007)

\bibitem{Iizuka2016Let}
Iizuka, S., Simo-Serra, E., Ishikawa, H.: Let there be color!: joint end-to-end
  learning of global and local image priors for automatic image colorization
  with simultaneous classification. ACM Trans. Graph.  35,  110:1--110:11
  (2016)

\bibitem{Isola2016Image}
Isola, P., Zhu, J.Y., Zhou, T., Efros, A.A.: Image-to-image translation with
  conditional adversarial networks. CoRR  abs/1611.07004 (2016)

\bibitem{Johnson2016Perceptual}
Johnson, J., Alahi, A., Fei-Fei, L.: Perceptual losses for real-time style
  transfer and super-resolution. In: ECCV (2016)

\bibitem{Kingma2014Auto}
Kingma, D.P., Welling, M.: Auto-encoding variational bayes. In: Conference
  proceedings: papers accepted to the International Conference on Learning
  Representations (ICLR) 2014 (2014)

\bibitem{Lee2009Convolutional}
Lee, H., Grosse, R., Ranganath, R., Ng, A.Y.: Convolutional deep belief
  networks for scalable unsupervised learning of hierarchical representations.
  In: International Conference on Machine Learning, ICML 2009, Montreal,
  Quebec, Canada,. pp. 609--616 (2009)

\bibitem{Mirza2014Conditional}
Mirza, M., Osindero, S.: Conditional generative adversarial nets. CoRR
  abs/1411.1784 (2014)

\bibitem{Pathak2016Context}
Pathak, D., Krahenbuhl, P., Donahue, J., Darrell, T., Efros, A.A.: Context
  encoders: Feature learning by inpainting. CVPR pp. 2536--2544 (2016)

\bibitem{Radford2015Unsupervised}
Radford, A., Metz, L., Chintala, S.: Unsupervised representation learning with
  deep convolutional generative adversarial networks. CoRR  abs/1511.06434
  (2015)

\bibitem{Reed2016Generative}
Reed, S., Akata, Z., Yan, X., Logeswaran, L., Schiele, B., Lee, H.: Generative
  adversarial text to image synthesis. In: International Conference on Machine
  Learning (2016)

\bibitem{Ronneberger2015U}
Ronneberger, O., Fischer, P., Brox, T.: U-Net: Convolutional Networks for
  Biomedical Image Segmentation (2015)

\bibitem{Russakovsky2015ImageNet}
Russakovsky, O., Deng, J., Su, H., Krause, J., Satheesh, S., Ma, S., Huang, Z.,
  Karpathy, A., Khosla, A., Bernstein, M.: Imagenet large scale visual
  recognition challenge. International Journal of Computer Vision  115(3),
  211--252 (2015)

\bibitem{Salakhutdinov2009Deep}
Salakhutdinov, R., Hinton, G.: Deep boltzmann machines. Journal of Machine
  Learning Research  5(2),  1967 -- 2006 (2009)

\bibitem{Sangkloy2016Scribbler}
Sangkloy, P., Lu, J., Fang, C., Yu, F., Hays, J.: Scribbler: Controlling deep
  image synthesis with sketch and color. CoRR  abs/1612.00835 (2016)

\bibitem{Simonyan2014Very}
Simonyan, K., Zisserman, A.: Very deep convolutional networks for large-scale
  image recognition. CoRR  abs/1409.1556 (2014)

\bibitem{Wang2016Generative}
Wang, X., Gupta, A.: Generative image modeling using style and structure
  adversarial networks. CoRR  abs/1603.05631 (2016)

\bibitem{Winnem2012XDoG}
Winnem{\"o}ller, H., Kyprianidis, J.E., Olsen, S.C.: Xdog: An extended
  difference-of-gaussians compendium including advanced image stylization.
  Computers {\&} Graphics  36,  740--753 (2012)

\bibitem{Zhang2016Colorful}
Zhang, R., Isola, P., Efros, A.A.: Colorful image colorization. CoRR
  abs/1603.08511 (2016)

\end{thebibliography}
\end{document}